\documentclass[smallextended]{svjour3} 
\smartqed

\begin{document}


\title{The Singleton Fallacy}
\subtitle{Why Current Critiques of Language Models Miss the Point}

\titlerunning{The Singleton Fallacy}

\author{Magnus Sahlgren \and
        Fredrik Carlsson}
        
\institute{RISE (Research Institutes of Sweden)\\
NLU Research Group\\
Sweden\\
\email{magnus.sahlgren@ri.se, fredrik.carlsson@ri.se}
}


\date{Received: date / Accepted: date}

\maketitle

\begin{abstract}
This paper discusses the current critique against neural network-based Natural Language Understanding (NLU) solutions known as {\em language models}. We argue that much of the current debate rests on an argumentation error that we will refer to as {\em the singleton fallacy}: the assumption that language, meaning, and understanding are single and uniform phenomena that are unobtainable by (current) language models. By contrast, we will argue that there are many different types of language use, meaning and understanding, and that (current) language models are build with the explicit purpose of acquiring and representing one type of structural understanding of language. We will argue that such structural understanding may cover several different modalities, and as such can handle several different types of meaning. Our position is that we currently see no theoretical reason why such structural knowledge would be insufficient to count  as ``real'' understanding.
\end{abstract}

\section{Introduction}
\label{intro}

We are at an inspiring stage in research on Natural Language Understanding (NLU), with the development of models that are capable of unprecedented progress across a wide range of tasks \cite{NEURIPS2019_4496bf24}. At the same time, there are critical studies being published that demonstrate limitations of our current solutions \cite{niven-kao-2019-probing,mccoy-etal-2019-right,ribeiro-etal-2020-beyond}, and more recently, voices have been raised calling for, if not taking a step back, then at least to stop for a moment and recollect our theoretical bearings \cite{bisk2020experience,bender-koller-2020-climbing}. Even if these latter theoretical contributions have slightly different perspectives --- \cite{bisk2020experience} introduce the notions of \textit{World Scopes} as a way to argue for the futility of using only text data to train NLU models, while \cite{bender-koller-2020-climbing} posit a strict distinction between form and meaning, arguing that models only trained on form cannot grasp meaning --- they share what we consider to be a healthy skepticism of the currently somewhat opportunistic and methodologically narrow-minded development.

The main controversy in this recent debate is the question to what extent our current NLU approaches --- i.e.~predominantly Transformer neural network language models --- can be said to really {\em understand} language, and whether the currently dominating research direction has any potential at all to lead to models with actual understanding. To put the point succinctly: will it eventually prove to be enough to train a thousand-layer trillion-parameter Transformer language model on the entire world's collected texts, or do we need something more or something else to reach true NLU (and what is ``true'' NLU anyway)? The recent excitement and hype in news and popular science press surrounding GPT-3 (see e.g.~\cite{ft:gpt3} and \cite{forbes:gpt3}) of course does nothing to dampen this controversy. 
While we share the overall assessment that more theoretical considerations would be beneficial for current and future NLU development, we think that both \cite{bisk2020experience} and \cite{bender-koller-2020-climbing} oversimplify important core discussion points. Our sentiment is hence that some of the presented arguments in the debate are somewhat misplaced and insufficient, as we find them to misrepresent the inherent complexity in discussing controversial topics such as meaning and understanding. This contribution therefore aims to analyze, and hopefully clarify, some of these arguments while also raising some novel discussion points of its own. 

\section{Language, Meaning and Understanding: a Philosophical Trifecta}
\label{lang:under}

It is always precarious to build arguments on inherently vague and general concepts such as ``language'', ``understanding'' and ``meaning,'' as the resulting theoretical constructs may become so overly general that they almost become vacuous. In this first section, we discuss how these terms are used in the current debate, and we argue that most of the current critique of the semantic capabilities of language models rest on a misunderstanding we refer to as {\em the singleton fallacy}. In short, this argumentation error consists in assuming that a term refers to a single uniform phenomenon, when in practice the term can refer to a large set of phenomena connected by family resemblances. Sections \ref{Language Is not a single thing} and \ref{What is understanding} discusses the concepts of ``language'' and ``understanding,'' while Section \ref{Current NLU} focuses on how current language models understand language.

\subsection{Language is Not One Single Thing}
\label{Language Is not a single thing}

Language is normally defined as the system of symbols that we use to communicate,
and learning a language entails learning the set of symbols and rules that define the system. Learning the set of symbols equals vocabulary acquisition, while learning the rules entails recognizing and formalizing grammatical, morphological, and syntactic regularities. We measure these competencies in humans --- often indirectly --- by using various language proficiency tests, such as vocabulary tests, cloze tests, reading comprehension, as well as various forms of production, interaction, and mediation tests (such as translation and interpretation). 
To evaluate our current NLU solutions, we often use specifically designed test sets, such as GLUE \cite{wang-etal-2018-glue}, SuperGLUE \cite{NEURIPS2019_4496bf24} and Dynabench \cite{nie-etal-2020-adversarial}, or more specific probing tasks that attempt to more directly measure a model's capacity to represent a specific linguistic phenomenon \cite{tenney2018what,liu-etal-2019-linguistic,jawahar-etal-2019-bert}. Even if current language models have been shown to underperform in some specific test settings (such as their ability to handle negation \cite{ettinger-2020-bert}), there is an overwhelming body of empirical results to demonstrate that current language models have a passable capacity to detect the symbols and rules that define human language. This is not what is under dispute in the current debate; what {\em is} under dispute is whether such structural knowledge about the workings of the language system suffices.

Of course, the question is: suffices for what? Presumably, we develop NLU systems in order to do the things we humans do with language. And here is the complication; we do not only do {\em one} thing with language. We humans do lots of different things with language, ranging from primal vocal expressions, over basic naming of objects and responding to simple questions, to more complicated tasks such as following instructions, arguing, or participating in negotiations. Language behavior is decidedly not one single activity, but a collection of many interrelated competencies and activities that together constitute the totality of human linguistic behavior. \cite{wittgenstein1953philosophical} refers to the relations between these interrelated linguistic activities as {\em family resemblances}, and he explains the situation thus: ``Instead of producing something common to all that we call language, I am saying that these phenomena have no one thing in common which makes us use the same word for all --- but that they are related to one another in many different ways. And it is because of this relationship, or these relationships, that we call them all `language'.'' 

All humans have a slightly different set of linguistic abilities, and even if two language users share a linguistic ability --- e.g.~arguing --- they are typically not equally good at it. Linguistic proficiency is a continuous scale, ranging from more or less complete incompetence to more or less complete mastery. We normally do not think about (human) language learning and linguistic proficiency as a pursuit of one single ultimate goal, whose completion is a binary outcome of either success or failure, but rather as a collection of tasks that can be performed in a number of different ways. When trying to define a criteria for quantifying linguistic proficiency, one must acknowledge that a totality of these linguistic skills is never actually manifested (as far as we know) in one single human language user. All current language learning tests are modelled according to this assumption, and hence deliver scalar results on a number of different test settings. 

We suggest that it may be more productive to explicitly recognize that language behavior covers a plethora of activities, and that linguistic competence can only be measured on a continuous scale. Doing so will also foster more realistic expectations on NLU solutions. Instead of demanding them to be flawless, generic and applicable to every possible use case, we may be better off adopting the same type of expectations as with human language users; they will all be different, and --- importantly --- good at different things.

\subsection{How Should We Understand ``Understanding''?}
\label{What is understanding}

The main controversy in the current debate is not so much whether language models can be trained to perform various types of language games (most commentators seem to agree that they indeed can). The main controversy is instead whether a language model that has been trained to perform some language game actually has any {\em real} understanding of language, or if it is confined to purely structural knowledge. In order to discuss this question seriously, we first need to understand what ``understanding'' means. \cite{bender-koller-2020-climbing} propose that understanding is pairing an expression with the correct meaning. We argue that this is a clear instance of the singleton fallacy (i.e.~assuming that understanding is one single thing), and it is certainly {\em not} consistent with how we use the term ``understanding'' in normal language use. Perhaps a couple of examples can be enlightening.

Let us begin by considering a type of expression that seems to be of major concern for \cite{bisk2020experience}: an instruction such as ``perform a backflip and land in a split.'' We can encounter a number of different types of criteria for understanding such an instruction in normal language use. One criterion might be to simply to be able to write an essay about acrobatics and use the expression consistently (and correctly) in text. Another is the ability to point to someone who actually performs a backflip and lands in a split, or to draw an image of someone performing the action. Yet another criterion is the ability to actually perform a backflip and land in a split, or even to just attempt and fail. On the other hand, if someone replies to the instruction with a profanity, or by simply leaving the room, we may be more hesitant to grant them the capacity of understanding. Unless, of course, we actually intended the expression to be an insult.

An instruction is admittedly an example of a relatively complicated language game. Let us for the sake of argument also consider one of the arguably simplest forms of linguistic expressions: a concrete noun such as ``door.'' When do we say that someone understands this simple word? Is it enough if they can use the word correctly in sentences, and exchange it for other similar words in context, or do they need to be able to visualize the object, and to be able to point out a correct object in the world? What about a situation when someone approaches a door carrying some heavy object, upon which the person looks at you and calls out ``door!'' Would it count as correct understanding in this case to reply ``yes, it is'' (and, optionally, to point at the door), or do you need to be able to infer that the person needs assistance with opening the door, and that you therefore should open the door for her? We would probably expect most human language users to open the door in response to such an utterance, but there will definitely be some variation of outcomes, with some people not being able to grasp the performative expectations of the speaker (but still having a perfectly good understanding of the term ``door'' and what it refers to), and some people perhaps instead offering to carry the load instead of opening the door (and thereby literally misinterpreting the speaker, while still performing a suitable action). 

The point of these somewhat wordy examples is to demonstrate that there can be many different types of (correct) understanding of a given linguistic expression. 
One type is the intra-linguistic, structural type of understanding that enables the subject to produce coherent linguistic output. Another is the referential understanding that enables the subject to identify (and visualize) corresponding things and situations in the world, a third is the social understanding that enables the subject to interpret other peoples intentions, and a possible fourth type is performative understanding where a person can (try to) accomplish the action being mentioned. These different types of understanding map approximately to \cite{bisk2020experience} World Scopes 1--3 (intra-linguistic understanding), 4 (referential understanding), and 5 (social understanding). 
In more traditional linguistic terms, we might use the terms {\em conventional} meaning, {\em referential} meaning, and {\em pragmatic} meaning to refer to these different types of information content. The definition of understanding proposed by \cite{bender-koller-2020-climbing} primarily applies to pragmatic aspects of language use, where the task of the interlocutor is to identify the intents of the speaker (or writer).

\subsection{The Structuralism of Contemporary NLU Approaches}
\label{Current NLU}
It may be useful at this point to consider how (one breed of) current NLU systems ``understand'' language. A particularly successful approach to NLU (and NLP in general) at the moment is to use deep Transformer networks that are trained on vast amounts of language data using a language modeling objective, and then specialized (or simply applied) to perform specific tasks \cite{collobert:weston,Peters:2018,devlin-etal-2019-bert,brown2020language}. Such models implement a fundamentally structuralist --- and even {\em distributionalist} \cite{Sahlgren2008,gastaldi}--- view on language, where a model of how symbols are combined and substituted for each other is learned by observing samples of language use. The language modeling component (or, in somewhat older methods, the embeddings) encodes basic knowledge about the {\em structure} of the language system, which can be {\em employed} for solving specific linguistic tasks. This is eminently well demonstrated in the recent work on zero-shot learning \cite{yinroth2019zeroshot,brown2020language}. 

The ``understanding'' these models have of language is entirely structural, and thus does not extend beyond the language system --- or, more accurately, beyond the structural properties of the input modality. Speaking in terms of the different types of understanding we discussed above, this refers to conventional, intra-linguistic understanding. Note that this is a very intentional restriction, since the learning objective of these models is optimized for learning distributional regularities. It follows that if the input signal consists of several different modalities (e.g.~language {\em and} vision, sound, touch, and maybe even smell and taste), then the resulting structural knowledge will cover {\em all} of these modalities. A distributional model built from multimodal data will thus be able to employ its (structural) knowledge cross-modally, so that, e.g., vision data can affect language knowledge and vice versa \cite{NEURIPS2019_4496bf24,Su2020VL-BERT:}. 
Such a multimodal model may be able to form images of input text, so that when given an input such as ``door'' it can produce, or at least pair the text with, an image of a door \cite{dall-e}. 
There have also been a fair amount of work on image captioning, where a model produces text based on an input image \cite{Herdade2019ImageCT,Li_2019_ICCV}.

This cross-modal ability cannot be described as purely intra-linguistic, since it covers several modalities. While we might not want to go as far as to call this referential understanding, it should certainly count as {\em visual} understanding (of language). It is an interesting question how we will view distributional models when we start to incorporate more modalities in their training data. What type of understanding would we say a language model has if it can connect linguistic expressions to actions or situational parameters? Multimodality is mentioned by both \cite{bisk2020experience} and \cite{bender-koller-2020-climbing} as a promising, if not necessary, research direction towards future NLU, and we agree; it seems like an unnecessary restriction to only focus on text data when there is such an abundance of other types of data available. However, there is also a more fundamental question in relation to multimodality, and that is whether there are things that {\em cannot} be learned by merely reading large bodies of text data?

\section{There is Nothing Outside the Text}

The last section problematized the use of philosophically nebulous terms such as ``understanding'' and ``meaning,'' and we argued that careless use of such terms invites to an argumentation error that we labelled the singleton fallacy. In this section, we discuss some of the more concrete arguments against language models, and we argue that they inevitably collapse into dualism, which we consider to be a defeatist position for an applied computational field of study. A consistent theme in the current critique of language models is the assumption that text data is insufficient as learning material in order to reach real understanding of language. We have already noted the perils of using such general statements as ``understanding of language,'' but in this section we will take a closer look at some of the specific arguments and thought experiments that motivate this assumption. 


\subsection{The Symbol Manipulators: Octopuses, Super-Scientists, and Language Models}

Both \cite{bisk2020experience} and \cite{bender-koller-2020-climbing} clearly think that text data is insufficient to reach true NLU. \cite{bisk2020experience} refer to the meaning of ``painting'' as a case in point when a purely linguistic signal will be insufficient to learn ``the meaning, method and implications'' of the concept. It is not clear to us that this is the case; learning from language {\em is} learning about our world; we use language to communicate and store experiences, opinion, facts and knowledge. 
\cite{bender-koller-2020-climbing} constructs a more elaborate thought experiment to make basically the same point. The thought experiment features a hyper-intelligent octopus (``O'') that inserts itself in the middle of a two-way human communication channel, and that eventually (due to loneliness, we are told) tries to pose as one of the human interlocutors. 

We are not convinced by the ``octopus test.'' A being that inserts itself into a communication channel because of loneliness obviously needs to have some sort of prior concept of, experience with, and desire to take part in, communication. 
But perhaps we should only assume that O coincidentally happened to connect to the communication channel, and that it as a reflex, or due to programming, start to detect and process statistical patterns in the symbol streams. In other words, O is a language model. We then agree that if the ``O model'' only learns from the information being transmitted between A and B, it will not be able to generalize outside the context of A and B. But what if O has also connected to the underwater transatlantic Internet cable, and has learned continuously from all the world's web traffic (let's restrict the learning material to text context for now) since the cable was submerged, some 20 years ago?\footnote{The latest transatlantic telecommunications cable (TAT-14) was installed 2000.} Would it then be completely unrealistic to think that O would actually be able to respond something reasonably creative given the question how to use sticks to defend against a bear?

\cite{bender-koller-2020-climbing} main point of the thought experiment is to argue that symbol manipulation by itself cannot transcend beyond the symbol system. This is certainly true from one perspective; the O language model, trained with the type of objective we currently use, would not be able to actually recognize a bear or a pair of sticks in the real world (since it ``lives'' in a purely textual world). On the other hand, it has certainly learned things {\em about} the world by reading the entire Internet, so the model might be able to produce accurate textual descriptions of both bears and sticks, based on information it has acquired from reading, but it will not be able to {\em act} outside the textual modality. This seems to be a main point for both \cite{bender-koller-2020-climbing} and \cite{bisk2020experience}, and we agree that a purely textual model will be constrained to a purely textual reality. 

This raises the question (again) whether there are any {\em a priori} constraints for what can and cannot be learned from language? Imagine, for the sake of argument, that we are a brain in a vat with super-human abilities enabling us to read, understand, and remember everything that ever has been written (and maybe even hear everything that has been spoken) about painting, bears, sticks, and animal attacks. Imagine further that we at some later stage would be implanted into a body; would we then learn anything new when we are actually able to perform the activities in question (e.g.~painting and using sticks to defend bear attacks)? We are at this point getting dangerously close to being sucked into the infamous ``Mary the super-scientist argument'' \cite{Ludlow2004-LUDTSA} with questions about the reality of subjective experience, and nebulous terms such as ``qualia.'' 

We suggest to stay well clear of this debate by using more precise terminology. If by ``understanding'' we refer to structural knowledge, then obviously language data will be enough; if we instead refer to the ability to perform actions in the world, then, equally obviously, we need to also acquire motor skills and connect them to linguistic expressions. We completely agree that no language model, no matter how much data it has been trained on and how many parameters it has, by itself will be able to understand instructions in the sense of actually performing actions in the world. But we {\em do} believe that, in principle, a language model can acquire a similar understanding and mastery {\em of language} as a human language user. 

\subsection{Communicative Intent and the Cartesian Theater}

Even if we disagree that the octopus test disproves that language models actually understand language, we {\em do} think it points to one important aspect of language use: namely that of {\em agency}. That is, if O is only a language model, we would not expect it to spontaneously reply to some statement from A or B, since it has no incentives to do so. A language model only produces an output when prompted; it has no will or intent of its own. Human language users, by contrast, have plans, ambitions and intents, which drive their linguistic actions. We humans play language games in order to achieve some goal, e.g.~to make someone open a door, or to insult someone by giving them an instruction they cannot complete. Language models (and other current NLU techniques) {\em execute} language games when prompted to do so, but the intent is typically supplied by the human operator.

Of course, there is one specific NLP application that is specifically concerned with intents: dialogue systems (or chatbots, to use the more popular vernacular). Consider a simple chatbot that operates after a given plan, for example to call a restaurant and book a table for dinner. Such a chatbot would not only be able to act according to its own intents, but would presumably also be able to recognize its interlocutor's intents (by simply classifying user responses according to a set of given intent categories). 
If we were to take the position that pragmatics is the necessary requirement for understanding, as \cite{bender-koller-2020-climbing} seems to do, it would lead to the slightly odd consequence that a simple (perhaps even completely scripted) dialogue system would count as having a fuller understanding of language than a language model that is capable of near-human performance on reading comprehension tasks. Such a comparison is of course nonsensical, since these systems are designed to perform different types of tasks with different linguistic requirements and thus cannot be compared on a single scale of understanding (there is no such thing).

This reductio ad absurdum example demonstrates the perils of strict definitions, such as the $M \subseteq E \times I$ formula for meaning suggested by \cite{bender-koller-2020-climbing}.\footnote{$M$ is meaning, $E$ is expression, and $I$ is intent.} The main problem with invoking intents in a strict mathematical definition is that the concept is extremely vague. In operationalizations of intent recognition (e.g.~in chatbots), we operate with a limited and predefined taxonomy of intents that are relevant to a specific use case, but in open language use it is less clear how to assign intents to expressions. At what granular level do intents reside --- is it at the word level, sentence level, or speaker turn (and how does that translate into text, where a turn sometimes is an entire novel)? And is the subject always in a privileged position to identify her intents? This is questioned in particular by postmodern critical theory \cite{pluckrose2020cynical}, and there are plenty of examples in the public debate where the speaker's interpretation of her utterance differs from that of commentators (``I did not mean it like that'' is not an uncommon expression). We can even sometimes find ourselves in the curious position of being simultaneously correct and incorrect about a speaker's intent, which is the case with our example with ``door!'' in the previous section (if I don't open the door and instead take the person's load, I would have misinterpreted the literal intention with the utterance, but still have correctly interpreted the person's need). This lends a certain hermeneutic flavor to the concept of intent, which makes it slightly inconvenient to use in mathematical formulas.

We believe that pragmatics is no less, and perhaps even more, a product of conventionalization processes in language use than other types of understanding. This may be an uncontroversial statement, but it points to a question that certainly is not, namely whether pragmatic understanding can be acquired by only observing the linguistic signal. \cite{bender-koller-2020-climbing} clearly think not, and they argue that grasping intents requires extralinguistic knowledge. For a simple case such as ``door!'', this may entail being able to recognize the object referred to, and possibly also knowledge about how it operates. For more abstract concepts, \cite{bender-koller-2020-climbing} claim the existence of ``abstract'' or ``hypothetical world[s] in the speaker's mind.'' There is an apparent risk that the invocation of minds at this point collapses into Cartesian materialism \cite{dennett:consciousness}, which constructs the mind as a kind of control room (also referred to as a ``Cartesian theater'') where the subject --- the {\em self} or {\em homunculus} --- observes, interprets, and controls the outside world. We are not sure what \cite{bender-koller-2020-climbing} position would be with respect to such a view, but it is easy to see how it would posit the intents with the homunculus, which would then use the linguistic generator to express its intents in the form of language --- which is, in our understanding, more or less exactly what \cite{bender-koller-2020-climbing} propose.

\subsection{How the Current Critique Rekindles Distributionalism}

The distinction between {\em form} (i.e.~the linguistic signal) and {\em meaning} (i.e.~the intent) is central to \cite{bender-koller-2020-climbing}, and they claim that a model (or more general, a subject) ``trained purely on form will not learn meaning''. But if meanings can have an effect on the form (which we assume everyone agrees on), then a model should at least be able to observe, and learn, these effects. The point here is that there needs to be an accessible ``linguistic correlate'' to whatever meaning process we wish to stipulate, since otherwise communication would not be possible. 
Thus, in the sense that intentions (meanings) have effects on the linguistic signal (form), it will be possible to learn these effects by simply observing the signal.
It is precisely this consideration that underlies the distributional approach to semantics. \cite{harris54} provides the most articulate formulation of this argument: ``As Leonard Bloomfield pointed out, it frequently happens that when we do not rest with the explanation that something is due to meaning, we discover that it has a formal regularity or ‘explanation.’ It may still be ‘due to meaning’ in one sense, but it accords with a distributional regularity.''

Somewhat ironically, \cite{bender-koller-2020-climbing} objections to distributional approaches in the form of language models --- that meaning is something unobtainable from simply observing the linguistic signal --- thus effectively brings us back to the original motivation for using distributional approaches in computational linguistics in the first place: if meanings really are unobtainable from the linguistic signal, then all we can do from the linguistic perspective is to describe the linguistic regularities that are manifestations of the external meanings. 

\section{At the End of the Ascent} 

In the last section, we argued that the currently dominating approach in NLU --- distributionally-based methods --- originate in a reaction against precisely the kind of dualism professed by \cite{bender-koller-2020-climbing}. But even if the motivation for the current main path in NLU thereby should be clear, the question still remains whether this current path is feasible in the long run, or whether it will eventually lead to a dead-end. This section discusses what the dead-end might look like, and what that means for the hill-climbing question.

\subsection{The Chinese Room and Philosophical Zombies}

\cite{bender-koller-2020-climbing} main concern seems to be that our current research direction in NLU will lead to something like a Chinese Room. The Chinese Room argument is one of the classical philosophical thought experiments, in which \cite{Searle80minds} invites us to imagine a container (such as a room) populated with a person who does not speak Chinese, but who has access to a set of (extensive) instructions for manipulating Chinese symbols, such that when given an input sequence of Chinese symbols, the person can consult the instructions and produce an output that for a Chinese speaker outside the room seems like a coherent response. In short, the Chinese Room is much like our current language models. The question is whether any real understanding takes place in the symbol manipulating process? 

We will not attempt to contribute any novel arguments to the vast literature that exists on the Chinese Room argument, but we will point to the counter-argument commonly known as the ``system reply'' \cite{Searle80minds}. This response notes that for the observer of the room (whether it is an actual room, a computer, or a human that has internalized all the instructions) it will seem {\em as if} there is understanding --- or at least language proficiency --- going on in the room.  Similarly, for the user of a future NLU system (and perhaps even for certain current users of large-scale language models), it may seem as if the system understands language, even if there is ``only'' a language model on the inside. We can of course always question whether there is any ``real'' understanding going on, but if the absence or presence of this ``real'' understanding has no effect on the behavior of the system, it will be a mere epiphenomenon that need not concern us. The attentive reader will notice that this is a variant of what is commonly referred to as the Systems Reply to the Chinese Room argument, and is essentially the same argument as the Philosophical Zombies (or Zimboes, as Dennett calls them \cite{dennett:zombies}) that behave exactly like human beings, except that they have no consciousness. Such a being would not be able to {\em really} ``want,'' ``believe,'' and ``mean'' things, but we would probably still be better off using these terms to explain their behavior. 

\subsection{Understanding and the Intentional Stance}

This is what Dennett refers to as ``Intentional Stance'' \cite{Dennett87}: we ascribe intentionality to a system (or more generally, entities) in order to explain and predict its behavior. For very basic entities, such as a piece of wood, it is normally sufficient to ascribe physical properties to it in order to explain its behavior (it has a certain size and weight, and will splinter if hit by a sharp object). For slightly more complex entities, such as a chainsaw, we also ascribe functions that explain its expected behavior (if we pull the starter cord, the chain will start revolving along the blade, and if we put it against a piece of wood, it will saw through the wood). For even more complex entities, such as animals and human beings, it is normally not enough with physical properties and functional features to explain and predict their behavior. We also need to invoke intentionality --- i.e.~mental capacities --- in order to fully describe them. Note that we occasionally do this also with inanimate objects, in particular when their behavior starts to deviate from the expected functions: ``the chainsaw doesn't want to start!'' We are in such cases not suggesting that the chainsaw suddenly has become conscious in the same way as humans are conscious; it is simply more convenient to adopt an intentional stance in the absence of simpler functional explanations. It would probably be possible to provide purely functional, perhaps even mechanistic explanations at some very basic neurophysiological level for every action that an animal or human makes, but it would be quite cumbersome. The intentional stance is by far the more effective perspective. 

Dennett's point is that consciousness is not an extra ingredient in addition to the complexity of a system: consciousness {\em is} the complexity of the system. Our point is that understanding is also not an extra ingredient of a symbol manipulation system: ``understanding'' is a term we use to describe the complexity of such a system. Let us take a well-defined and clearly delimited possible use case for future androids as an example; a robotic doorman. Imagine that our doorman is equipped with a future version of a pre-trained Transformer language model trained on multimodal data (e.g.~both text and images) that has been fine-tuned on a relevant intent recognition task, and implemented in a dialogue engine with an interaction agenda (e.g.~to greet, help, and make small talk with the building's residents). 
Our doorman would probably have a good grasp of conventional as well as referential (or at least visual) meaning, and it would be able to both recognize (a limited set of) intents and act according to its own (limited set of) intents. 
Imagine further that our doorman is capable of performing a limited set of actions, such as opening the door, carrying a package, and calling for the elevator. In its very limited domain, our android would be more or less functionally indistinguishable from a human doorman. If someone were to approach the entrance carrying a stack of heavy boxes, and shouting ``door!'', our doorman would presumably be able to use the visual cues of the boxes to map the utterance to an intent category such as ``request for assistance'' rather than ``ostensive definition,'' and would thus be able to select a proper set of actions, such as opening the door or offering to carry the boxes. It seems strange to not use the explanation ``the android understood the utterance'' in such a case, even if it would in theory be possible to provide a complete functional description of the activations in the doorman's internal model. From our perspective, it would be much stranger to claim that the doorman did not actually understand the utterance, since it lacks a mapping from form to meaning.

\subsection{Montology}
Our point is that when the behavior of an NLU system becomes sufficiently complex, it will be easier to explain its behavior using intentional terms such as ``understanding,'' than to use a purely functional explanation. We posit that we are not far from a future where we habitually will say that machines and computer systems understand and misunderstand us, and that they have intents, wishes, and even feelings. This does not necessarily mean that they have {\em the same} type of understanding, intents, opinions, and feelings as humans do, but that their behavior will be best explained by using such terms. The same situation applies to animals (at least for certain people), and maybe even to plants (with the same caveat). We argue that the question whether there really is understanding going on, i.e.~whether there is also some mapping process executed in addition to the language use or behavior, is redundant in most situations. We can probably look forward to interesting and challenging ethical and philosophical discussions about such matters in the future, but for most practical purposes, it will be of neither interest nor consequence to question whether an NLU system is a mere symbol manipulator or a ``true'' understander.

It is important to understand which hill we are currently climbing, and why. As we have argued in this paper, the current hill, on which language models and most current NLU approaches live, is based on distributional sediment, which has amassed as a reaction to a dualistic view that posits understanding and meaning in a mental realm outside of language. The purpose of this hill is not to replicate a human in silico, but to devise computational systems that can manipulate linguistic symbols in a manner similar to humans. This is not the only hill in the NLU landscape --- there are hills based on logical formalisms, and hills based on knowledge engineering --- but based on the empirical evidence we currently have, the distributional hill has so far proven to be the incomparably most accessible ascent. 

It should thus come as no surprise that our answer to the question whether we are climbing the right hill is a resounding {\em yes}. We do, however, agree that we should not only climb the most accessible hills, and that we as a field need to encourage and make space for alternative and complementary approaches that may not have come as far yet. The most likely architecture for future NLU solutions will be a combination of different techniques, originating from different hills of the NLU landscape.

\section{Conclusion}

This paper has argued that much of the current debate on language models rests on what we have referred to as the singleton fallacy: the assumption that language, meaning, and understanding are single and uniform phenomena that are unobtainable by (current) language models. By contrast, we have argued that there are many different types of language use, meaning and understanding, and that (current) language models are build with the explicit purpose of acquiring and representing one type of structural understanding of language. We have argued that such structural understanding may cover several different modalities, and as such can handle several different types of meaning. Importantly, we see no theoretical reason why such structural knowledge would be insufficient to count as understanding. On the contrary, we believe that as our language models and NLU systems become simultaneously more proficient and more complex, users will have no choice but to adopt an intentional stance to these systems, upon which the question whether there is any ``true'' understanding in these systems becomes redundant. 


We are well aware that the current debate is of mainly philosophical interest, and that the practical relevance of this discussion is small to non-existent. A concrete suggestion to move the discussion forward is to think of ways to verify or falsify the opposing positions. We distinctively feel that the burden of proof lies with the opposition in this case; are there things we cannot do with language unless we have ``real'' (as opposed to structural) understanding? Which criteria should we use in order to certify NLU solutions as being ``understanders'' rather than mere symbol manipulators? Being able to solve our current General Language Understanding Evaluation benchmarks (i.e.~GLUE and SuperGLUE) obviously does not seem to be enough for the critics.

From the ever-growing literature on ``BERTology'' \cite{bertology} we know that there are tasks and linguistic phenomena that current language models handle badly, if at all, and we also know that they sometimes ``cheat'' when solving certain types of tasks \cite{niven-kao-2019-probing}. These are extremely valuable results, which will further the development of language models and other types of NLU solutions. However, these failures do not mean that language models are theoretically incapable of handling these tasks; it only means that our {\em current} models (i.e.~current training objectives, architectures, parameter settings, etc.) are incapable of recognizing certain phenomena. Which, we might add, is to be expected, given the comparably simple training objectives we currently use.

\bibliographystyle{spmpsci}
\bibliography{coling2020}

\end{document}